\documentclass[referee,sn-apa]{sn-jnl}
\usepackage[utf8]{inputenc}

\usepackage{pgf} 

\usepackage{graphicx}
\usepackage{amsmath}
\usepackage{amssymb}
\usepackage{bm}
\usepackage{listings}
\usepackage{wrapfig}
\usepackage{booktabs}
\usepackage{threeparttable}
\usepackage{bigstrut}
\usepackage{rotating,caption}
\usepackage{tgpagella}
\definecolor{royalblue(web)}{rgb}{0.25, 0.41, 0.88}

\usepackage{tabularx}
\usepackage{xltabular}

\usepackage{float}
\usepackage{listings}

\usepackage{quotes}

\newfloat{lstfloat}{htbp}{lop}
\floatname{lstfloat}{Code example}

\usepackage{multirow}

\usepackage{fontawesome5}

\usepackage{natbib}

\begin{document}

\title{Mapping Seven Decades of Philosophy in Colombia: Dynamic Topic Modeling of \emph{Ideas y Valores}}
\author*[1]{Juan R. Loaiza}
\affil*[1]{Departamento de Filosofía, Universidad Alberto Hurtado} 
\email{jloaiza@uahurtado.cl}
\author[2]{Miguel González-Duque}
\affil[2]{Análisis Cuantitativo}
\email{miguelgondu@gmail.com}
\date{}

\abstract{Data-driven approaches to philosophy have emerged as a valuable tool for studying the history of the discipline. However, most studies in this area have focused on a limited number of journals from specific regions and subfields. We expand the scope of this research by applying dynamic topic modeling techniques to explore the history of philosophy in Colombia and Latin America. Our study examines the Colombian philosophy journal \emph{Ideas y Valores}, founded in 1951 and currently one of the most influential academic philosophy journals in the region. By analyzing the evolution of topics across the journal's history, we identify various trends and specific dynamics in philosophical discourse within the Colombian and Latin American context. Our findings reveal that the most prominent topics are value theory (including ethics, political philosophy, and aesthetics), epistemology, and the philosophy of science. We also trace the evolution of articles focused on interpreting a specific philosopher's work rather than proposing new positions, and we note a salient emphasis on German philosophers such as Kant, Husserl, and Hegel across various topics throughout the journal's lifetime. Given the journal's founding aspiration toward more original, non-exegetical philosophy, we investigate whether exegetical topics became comparatively less prominent over time. Our analysis suggests no significant decline in such topics. Finally, we propose ideas for extending this research to other Latin American journals and suggest improvements for natural language processing workflows in non-English languages.

}

\maketitle

\section{Introduction}

Digital analyses and tools have begun impacting philosophy in recent years, opening a number of questions and possibilities for researchers in philosophy and the digital humanities to explore. Some of these recent contributions include analyzing philosophical journals to reconstruct the history of certain debates in academic philosophy. One such study \citep{MalaterrePoS} analyzed the journal \emph{Philosophy of Science} and studied how the philosophy of science as a discipline has evolved from 1934 to 2015. Among its most interesting findings we can see how philosophy of science shifted from debates on deductive-nomological explanation to mechanistic explanations, how metaphysical and ontological problems became more present in recent decades, and how trends in philosophy of biology and philosophy of medicine have emerged in the last years. Later, \citet{MalaterreB&P} applied a similar method to the journal \emph{Biology and Philosophy}, and \citet{Malaterre2021} compared several journals in philosophy of science.

Given the potential of using digital techniques to analyze the history of philosophy, we set out to apply these strategies to another particular case. In this study, we aim to describe the development of academic philosophy in Colombia from 1951 to 2022 using text mining techniques on a well-known philosophy journal in Colombia: \emph{Ideas y Valores}. Based at the Universidad Nacional de Colombia, the most important public institution in the country, \emph{Ideas y Valores} was founded in 1951. The journal has been one of the main outlets for Colombian as well as Latin American philosophers more generally, publishing to date more than 1,000 documents including original articles, reviews, and other document types. Given its impact in the country and in the region, we propose to take \emph{Ideas y Valores} as a lens through which we can observe how philosophical ideas were developed and settled in Colombian (and to some extent Latin American) academic philosophy throughout the journal's lifetime.

The following study inquires into which topics are more prominent throughout the history of \emph{Ideas y Valores} and how they have evolved through time. Additionally, we ask whether articles focusing on exegetical topics have decreased as years pass, given the academic context in which the journal started, its editorial policies, and recent trends in academic publishing in philosophy, the relevance of which will be discussed below. To answer these questions, we applied topic modeling techniques to the journal's original articles using Dynamic Topic Modeling \citep{BleiDTM} to construct a timeline of the evolution of specific topics from 1951 to January 2022.

Our findings show that \emph{Ideas y Valores} has focused largely on topics related to value theory (ethics, political philosophy, and aesthetics) and epistemology and philosophy of science. We also see the impact of German philosophy in different areas of academic philosophy in the journal's publications, with various topics relating directly to the works of Kant, Husserl, Heidegger, and Hegel, among other German thinkers. Lastly, we evaluate whether exegetical topics have decreased in focus throughout the years in favor of more contemporary and argumentative (in contrast to interpretative) literature. These results suggest that no such change has occurred, and that topics with a historical focus remain as present as they have been since the journal's foundation. Overall, these findings offer a snapshot of the history of academic philosophy in Colombia and Latin America, and they invite further research into the trends and distribution of topics in philosophy in the region.

\section{Background}

\subsection{History of philosophy in \textit{Ideas y Valores}}

\label{sec:history_of_the_journal}

\emph{Ideas y Valores} was founded in 1951 by Cayetano Betancur, director of the \emph{Instituto de Filosofía} at the Universidad Nacional de Colombia. The project was to found a journal that would offer a space of discussion for all sorts of ideas, including literary texts. The journal published eight issues between 1951 and 1954, before entering a hiatus that lasted until 1962. From then on, the journal has published at least one issue almost every year, with a couple of minor exceptions. From 1983 onwards, the journal has been published continuously until today. As of the time of writing, \emph{Ideas y Valores} is a well-known journal in Colombia and Latin America, indexed in Scopus and Web of Science, with all of its publications available in Gold Open Access under a Creative Commons BY-NC-ND 4.0 license. According to the \citet{scimago} Journal Rankings in 2024 (i.e., ranking 2023), it has an h-index of 7, which, while being low for philosophy internationally, is still among the highest for philosophy journals in Latin America (which have an average h-index of 4).

\emph{Ideas y Valores} has undergone several changes in publication policy throughout its years. The journal was founded during a period where philosophical reflection occurred, not only in the form of original philosophical articles (as we would find them in most journals nowadays), but also literary works, translations of foreign texts into Spanish (including works by Hegel, Heidegger, and others), political commentary, and the communication of philosophical ideas to the general public. According to various historians of philosophy in Colombia \citep{Rios2020,Betancur2015a,Tovar2000}, even though philosophical thought in Colombia had been developed before the foundation of institutions like the Facultad de Filosofía y Letras (the Faculty of Philosophy and Literature) at Universidad Nacional de Colombia in 1946, such thought was mostly carried out in these non-academic contexts. As such, the first years of the corpus are very heterogeneous.

In this historical context, the early editors' aspirations differed from this more cultural, political, or even religious picture of philosophical thought that dominated the Colombian philosophical landscape before the 1950s \citep{SierraMejia1985,Gomez2021}. Due to the non-institutional nature of these works, a group of Colombian philosophers, including names like Cayetano Betancur (founder and first editor of \emph{Ideas y Valores}), Francisco Romero, Rafael Carrillo, and Rubén Sierra Mejía, did not consider these ideas proper philosophy \citep{IsazaDuque2013}. As part of their project of consolidating a more institutionalized and academic philosophy in Colombia, they founded \emph{Ideas y Valores}. The journal was a strategy to promote a philosophical style that was geared toward discussing and proposing original views that applied elsewhere as much as in the local context, contrasting with the ``scholastics'' that focused exclusively on presenting others' thought \citep{SierraMejia1967} and with work assumed to be more literary or politically activist rather than academic \citep{IsazaDuque2013}. Early reports on the reception of the journal in Colombia and abroad celebrated this policy and highlighted the journal as a venue of original work instead of ``repeated studies'' or literary, aesthetic, or rhetorical views \citep{Recepcion1952,Recepcion1953}. This editorial policy has characterized the journal ever since, and has rippled to this day in Colombian philosophy's historiography \citep{Rios2020}.

The Colombian philosophical landscape has been influenced by different traditions and schools since the journal's inception. Before the 1990s, the most prominent tradition was arguably the phenomenological tradition of Husserl and Heidegger. Yet, during the 1990s other traditions flourished, including the postmodern tradition inspired by French thinkers such as Deleuze and Foucault \citep{Soto2017}, the analytic tradition focused on logic and language (especially in the philosophy of Wittgenstein) \citep{IsazaDuque2013}, and later on, on problems related to the philosophy of mind \citep{Perez2023}. Even though there is an appreciation of these shifts among historians of philosophy and academic philosophers in Colombia and Latin America, it is relatively unclear how these authors, traditions, topics, and ideas have shifted. This motivates a more data-driven approach to model how these ideas have changed through time.

\subsection{Dynamic Topic Models}
\label{subsec:topic_modelling_using_dtms}

\begin{figure*}[t]
    \centering
    \caption{Words, topics, and documents in a dynamic topic model. Original articles (a) from \textit{Ideas y Valores} are processed into bag-of-words by extracting and fixing the text from either PDF or HTML sources, followed by a normalization process in which stop-words are removed and words are lemmatized (b). Using our amassed collection of articles processed into bag-of-words, we use Dynamic Topic modeling to model each document as having a proportion of 90 different topics, 8 of which are shown in (c). By topic, we refer to a distribution over words that evolves over time. In the topic \textit{Wittgenstein}, the words \textit{Wittgenstein}, \textit{lenguaje} (language), \textit{regla} (rule), and \textit{juego} (game) were the most probable from 2010 onwards, with slight variations in their likelihood in previous years (d). Dynamic Topic modeling accounts for how the discussion of a certain subject evolves over time.}
    \label{fig:dtm-explainer}
    \includegraphics[width=1.0\textwidth]{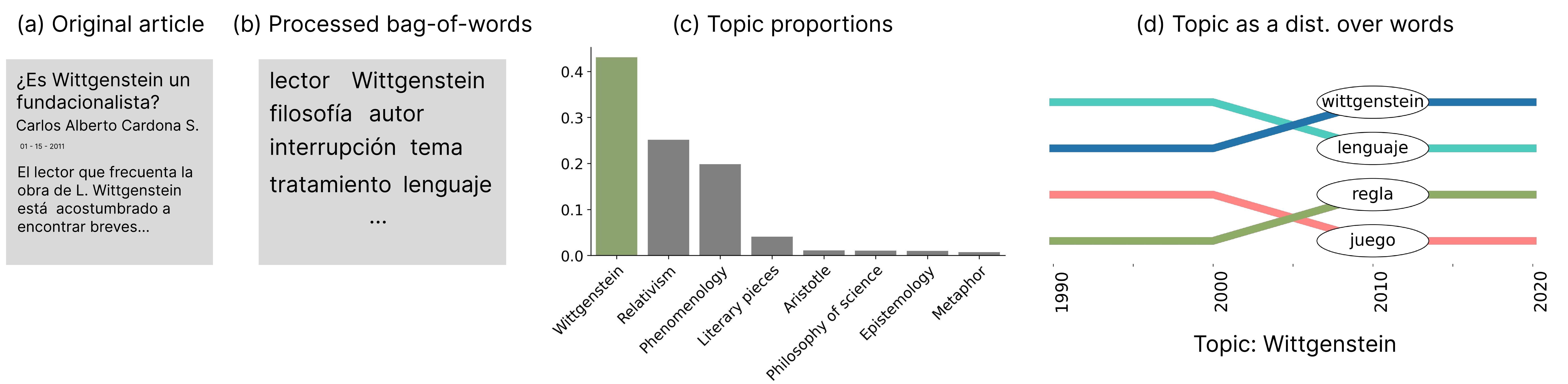}
\end{figure*}

In this analysis, we studied the corpus of \textit{Ideas y Valores}'s articles using Dynamic Topic Models (DTM; see \citet{BleiDTM}), which is an extension of the Latent Dirichlet Allocation model (LDA) proposed by \citet{BleiLDA}. These two models are examples of \textbf{topic modeling}, in which ``topics'' (broadly defined as collections of words that tend to co-occur) are automatically segmented. Both LDAs and DTMs are examples of \textbf{generative models}. Generative models describe, in probabilistic terms, the process by which text, images, or other modalities are created. In our setting, LDAs imagine a document from \textit{Ideas y Valores} as being made of a collection of \textit{topics} in certain proportions: an author might choose to write a paper that is 42\% about Wittgenstein, 25\% about relativism, and 20\% about phenomenology, and so on. Each of these topics (say, Wittgenstein) is made of words in different proportions (10\% ``Wittgenstein'', 8\% ``language'', 5\% ``rule'', etc.).\footnote{In mathematical terms, we would say that a document is a probability distribution over topics, and each topic is a probability distribution over words.} Fig.~\ref{fig:dtm-explainer} illustrates this relationship between words, documents, and topics using a specific article from our corpus.

LDAs then imagine the generative process as follows: once an author has settled on a group of topics and their proportions, the author writes each word of the document by (1) choosing a topic according to their proportions (e.g., choosing Wittgenstein with a probability of 42\% in our running example), and (2) selecting a word from said topic according to its probability (``rule'' with 5\% probability, and so on).

Fitting an LDA consists of determining these proportions from the corpus itself, using the tools of statistical inference. Once a number of topics $K$ is provided, LDA software determines which words belong to which topics, and in what proportion. The underlying mathematical process can be thought of as a form of ``clustering'': words that tend to happen together are clustered into the same topic, and how frequently they co-occur determines the proportion.

DTMs, the statistical model we use to uncover these latent topics, are an extension of LDAs that keep track of the temporal evolution of topics. Intuitively, DTMs learn how a topic transitions from one time period to the next. This transition takes the current topics and the proportions of the words therein, and slightly adjusts them to the topics discussed in the next time slice's documents. More formally, DTMs fit an LDA per time slice in the manner described above, sharing the topics across time slices using slight perturbations subsequently. That way, a topic is allowed to evolve over time: the words used for describing Wittgenstein might vary from decade to decade, but the underlying topic is still the same (see Fig.~\ref{fig:dtm-explainer}).

LDAs and DTMs are by no means the only forms of topic modeling. Other methods include Latent Semantic Indexing (LSI), Non-negative matrix factorization (NMF, recently used in the digital humanities context \citep[e.g.][]{Greene2024}), and more contemporary methods like BERTopic based on artificial neural networks and transformers \citep{bertopic:grootendorst:2022}.

\section{Methods}

\begin{figure*}[t]
    \centering        
    \caption{Number of articles and average length of documents per period of 5 years. Document length is computed after stopwords have been removed. The number of articles per 5-year period is represented by each bar against the left-hand y-axis; their average non-stopword length is represented by the red line and the right-hand y-axis.}\label{fig:num_articles}
    \includegraphics[width=\textwidth]{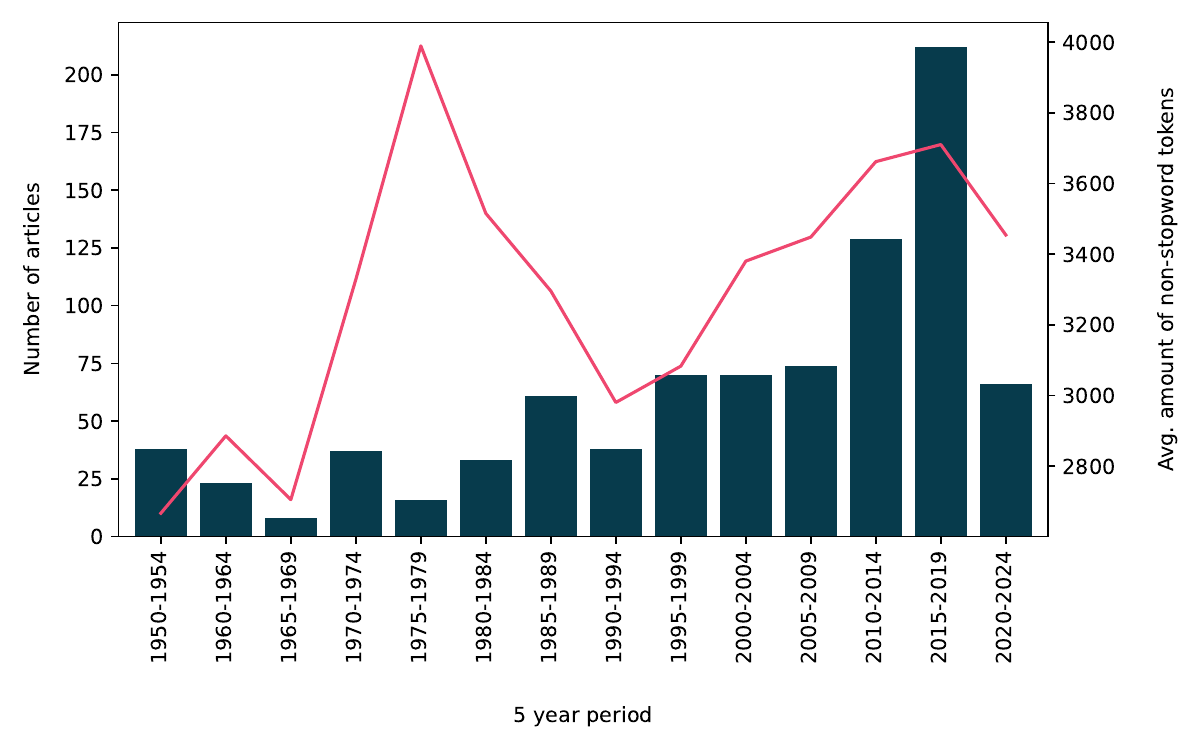}
\end{figure*}

For the present study, we retrieved all original articles in \emph{Ideas y Valores} available online and published between 1951 and January 2022. These articles are published in three languages: Spanish, English, and Portuguese. We analyze only those articles published in Spanish since they are the majority of the articles in the corpus (they constitute approximately 96\% of the total documents, with 1,615 out of 1,677 documents reported as in Spanish based on their metadata), and because natural language processing pipelines in topic modeling are not well-suited to analyze topics across different languages. Table \ref{tab:corpus_description} summarizes the resulting corpus. The corpus consisted of 875 full-text articles with an average word count of 3,429 words (Fig.~\ref{fig:num_articles}). We used articles in HTML format whenever possible, otherwise using articles in PDF format. This amounted to 275 articles in HTML format (particularly those between 2009 and 2017) and 600 articles in PDF format. For each of these formats, we used slightly different preprocessing procedures to account for various kinds of format-dependent artifacts as described in Sec.~\ref{subsec:data_preprocessing}. In addition to acquiring the articles' full text, we also acquired each article's metadata to aid in further analyses.

\subsection{Data acquisition and preprocessing}
\label{subsec:data_preprocessing}

To begin our study, we downloaded articles from \emph{Ideas y Valores}'s website. These articles are published under a Creative Commons Attribution-NonCommercial-NoDerivatives 4.0 International License, and are available either in PDF or HTML format. As explained above, we used articles in HTML format whenever possible, since they are easier to parse and analyze. Additionally, we obtained each article's metadata, which is available in each article's HTML headers, and saved it in JSON format to use for reference.

Data preprocessing proceeded in four steps: (1) text extraction, (2) punctuation and artifact removal, (3) stopword removal, and (4) lemmatization. The software implementation of these steps can be found in our GitHub repository (see \nameref{sec:data_availability}).

\begin{table}[t]
\centering
\caption{Counts of various elements in the entire \emph{Ideas y Valores} corpus from 1951 to January 2022.}
\label{tab:corpus_description}

\begin{tabular}{ll}
\toprule
{} &    Count \\ \midrule
Number of documents       &      875 \\
Number of words           &  3,001,085 \\
Number of unique words    &   138,192 \\
Number of corrected words &   438,834 \\
Average document length   &     3,429 \\
Number of stopwords       &     1,608 \\
Protected words           &       50 \\  \bottomrule
\end{tabular}
\vspace*{0.5em}
\footnotetext{Note: Words are tokens remaining after removing punctuation and tokens fewer than 3 characters long. Corrected words are token types that have been changed at least once during orthographical correction. Average document length is counted as the number of words remaining after removing stopwords and correcting the texts' orthography.}
\end{table}

\begin{figure}[t]
    \centering
\includegraphics[width=0.8\textwidth]{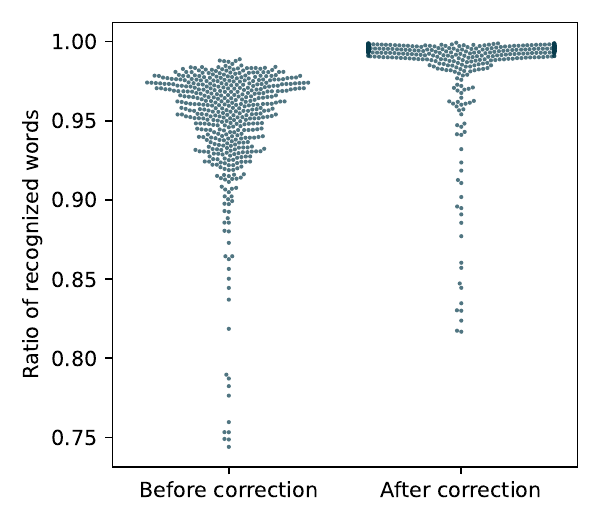}
\caption{Ratio of words recognized by PySpellChecker before and after orthographical correction. A word is recognized if it is contained in PySpellChecker's default Spanish, English, or German dictionaries, RAE's 2020 frequency list, or if it is contained in articles that were available in HTML format. Each point represents a single document, and its location represents how many words are recognized before and after correction against the whole document's word length.}
\label{fig:correction}
\end{figure}

For articles in HTML format, text extraction was straightforward, as the text is already in a format that is easy to process. We extracted the text from HTML files using the BeautifulSoup4 \citep{bs4} package version 4.9.3. For articles in PDF format, text extraction was slightly more difficult. Most PDF files had optical character recognition (OCR) data embedded, but such data often includes various artifacts (e.g., the letter ``M'' is often interpreted by OCR processes as the sequence ``\texttt{lVl}''). Hence, we first extracted all text data from each PDF file using a Python interface for Ghostscript \citep{Goebel2018}. Then we passed the text through an orthographic correction process using PySpellChecker \citep{Barrus2020}. This process used a list of word frequencies in Spanish collected by the \citet{RAECorpus} in order to attempt to correct each unknown word in a given text. Additionally, we implemented a custom dictionary using the HTML articles we had processed to obtain a more technical (i.e., philosophical) dictionary that could aid in correcting some terms. In total, we corrected 438,834 tokens (see Table \ref{tab:corpus_description}), improving how much of the text was recognizable by the dictionary on average from 94.97\% to 98.66\%. For the total corpus, word recognition improved from 95.3\% to 98.7\%. Figure \ref{fig:correction} shows the distribution of ratios of recognized words per article before and after correcting these tokens.

Punctuation and artifact removal in both cases was implemented using Python's native \verb|re| package. For further preprocessing steps for topic modeling, it is common practice to use libraries like the Natural Language Toolkit \citep[NLTK; ][]{NLTK} or spaCy \citep{spacy} for identifying stopwords, lemmatizing, and stemming \citep{igual_basics_2024}. For stopword removal we used three stopword lists (Spanish, English, and Portuguese) from the Natural Language Toolkit (NLTK) by \citet{NLTK}. Due to the limited number of words in the Spanish list, we added the list compiled by \citet{Diaz2016} and filtered it to protect philosophically important words such as ``verdad'' (truth), ``bien'' (good), and ``deber'' (duty). We then implemented both stopword removal and lemmatization using spaCy \citep{spacy}. We used spaCy instead of NLTK's own lemmatizer capabilities since NLTK's lemmatizer is much more limited for Spanish than spaCy's. For spaCy we used its own \verb|es_core_news_md| model, a model constructed from newspaper articles in Spanish.  After stopword removal and lemmatization we obtained a corpus of 875 documents with an average usable word count of 3,429 words. All lists and dictionaries can be downloaded from the study's GitHub repository (see \nameref{sec:data_availability}). 

\subsection{Data modeling}

As explained above, we used DTMs, which are an extension of LDA models (see Sec.~\ref{subsec:topic_modelling_using_dtms}), to extract topics from the corpus and study their evolution through time. Various implementations for LDA models can be found in the Python open-source community. In particular, the Python package \texttt{Gensim} \citep{gensim} provides a simple interface for computing topic models using not only LDA, but also other methods like LSI or NMF. \texttt{Gensim} also provides an interface for computing dynamic topic models, which is implemented in pure Python. Yet, we found \texttt{Gensim}'s implementation of DTMs to be slow in practice, and decided to use a C++ implementation provided by Blei's laboratory \citep{BleiLabExe}.

Using the DTM tooling provided by \citet{BleiLabExe}, we modeled the topics in our corpus using the following steps. First, recall that to be able to train a DTM, one needs to specify the number of topics $K$ beforehand. To decide on the number of topics, we inspected models in a grid $K\in\{50, 60, \dots, 150\}$, training each model using 3 different random seeds to account for the inherent stochasticity during training.

To decide which of these models to use, metrics like the average topic coherence, perplexity, number of topics without any articles assigned to them, and number of articles without  a topic were computed, following standard practice for topic models \citep[see e.g.,][]{Malaterre2021,Greene2024}. These metrics, combined with manual inspection of the models, led us to choose $K=90$ as the number of topics.\footnote{The exact metrics and their values can be found in our GitHub repository, alongside the code necessary to replicate this grid search. See \nameref{sec:data_availability}.}


\subsection{Topic interpretation}

Interpreting LDA models requires making the perspective of the researcher explicit. While we opted for a more data-driven approach to selecting the number of topics for our model (using coherence, log perplexity, number of topics without articles, and number of articles without a topic), topic interpretation requires a human (ideally expert) standpoint from which to judge the semantic content of each distribution. 

Topic interpretation was done in two phases. First, we manually inspected every topic, examining the most probable words and the most probable articles to appear according to the model. With LDA modeling techniques, each topic is a probability distribution over words, but it is the researcher who must give meaning to such distributions. This introduces a possible bias when it comes to analyzing such overarching corpora as the one we analyzed. Researchers are seldom equally knowledgeable in all areas of a given discipline. Rather, each researcher has a set of areas of expertise in which they are more knowledgeable, but as a result, researchers tend to be more sensitive to how their areas of expertise are subdivided than to how other areas are. For example, some observers might be more sensitive to differences in subdivisions within philosophy of mind or philosophy of science, but they might omit nuances when it comes to political philosophy or ethics. Similarly, readers in these areas may be sensitive to the various subareas within the latter, but they might be less sensitive to the heterogeneity within other areas of philosophy. 

To reduce such biases, given our areas of expertise, we opted to use an external categorization of academic philosophy as a reference to label each topic. Specifically, we decided to use the category taxonomy used in the popular PhilPapers \citep{PhilPapers} archive to guide labeling decisions and help reduce possible biases concerning the granularity of various categories. This taxonomy includes seven main areas (``Metaphysics and Epistemology'', ``Value Theory'', ``Science, Logic, and Mathematics'', ``History of Western Philosophy'', ``Philosophical Traditions'', ``Philosophy, Misc.,'' and ``Other Academic Areas''). Within each main area, there are various subdivisions of each field (e.g., ``Normative ethics'', ``Philosophy of Mathematics'', ``Philosophy of language'', etc.) which may contain further subdivisions. To label each topic, first we assigned a ``main area'' label to each topic, corresponding to one of five main areas in the PhilPapers taxonomy. These are ``Metaphysics and Epistemology'', ``Value Theory'', ``Science, Logic, and Mathematics'', ``History of Western Philosophy'', and ``Philosophical Traditions''. We omit ``Philosophy, Misc.'' and ``Other Academic Areas'' in the PhilPapers taxonomy because they are not meaningful for our analysis, but we do include an ``Other'' main area label for documents left in the corpus that are not academic articles (e.g., literary pieces published in the 1970s). After we assigned a main area to each topic, we assigned tags of different subareas inside that main area. These are often areas in the second level of the PhilPapers taxonomy (e.g., ``Epistemology'' is contained directly inside ``Metaphysics and Epistemology''), although in some cases we opt for more specific tags that could be informative.

Although this taxonomy proved useful, we acknowledge various drawbacks of this methodological decision to rely on PhilPapers' classification. First, this taxonomy is constructed with more attention to topics popular in contemporary anglophone philosophy, and in this regard it relegates certain topics to its margins. Second, it is unbalanced in certain areas, such as the history of philosophy or various strands of phenomenology and hermeneutics, which are lumped together in broad areas (``History of Western Philosophy'' and ``Philosophical Traditions'', respectively), in contrast to more granular distinctions in other fields in ``Science, Logic, and Mathematics'' and ``Metaphysics and Epistemology.'' Yet, there are several reasons why we used this taxonomy, despite its drawbacks. First, having a predefined, relatively standardized taxonomy to aid in tagging reduced such biases since it is easier to inspect which of the PhilPapers' categories would be a better fit for a given topic. This taxonomy includes divisions and subdivisions in most areas of philosophy, providing enough guidance to identify topics at various degrees of granularity. Second, although this taxonomy may also involve its editors' biases, we believe it at least provides an explicit and publicly available taxonomy that can be openly discussed in future research. Other candidate taxonomies will include biases based on their academic and historical context, and we therefore opt for a publicly available and open alternative rather than searching for an allegedly neutral classification. Lastly, the taxonomy includes topics from a wide range of historical periods, with various categories for topics discussed in more recent years, which fits the journal's lifespan and editorial policy that promoted working on topics of international (i.e., ``universal'') interest. In the last section, we will discuss these limitations and suggestions for future research.

Besides labeling each topic according to a main area and subareas, we also labeled topics based on their exegetical emphasis. These topics included works that focused on the interpretation of a particular piece of work or philosopher. We identified these topics based on the articles most likely to contain said topics. When these articles contained titles and abstracts focusing on interpreting a philosophical concept, particularly in the specific context of a philosopher's work (e.g., ``freedom'' in Spinoza, ``duty'' in Kant), we considered it an exegetical topic. We additionally inspected these topics to confirm that their most likely documents were not contemporary with the figures they were discussing, as could be the case with philosophers who were still alive and active during the journal's lifespan (e.g., Deleuze, Gadamer, among others). In these cases, we added a special tag ``\texttt{\#exegetical},'' which enabled us to examine the hypothesis that as the journal's aspiration to move away from ``scholastics'' \citep{SierraMejia1967} consolidated, articles shifted toward more contemporary discussions in academic philosophy.

\subsection{Assigning documents to topics}
\label{subsec:assigning_docs_to_topics}

As we mentioned above, the goal of topic modeling techniques is to compute a probability distribution over topics for each document. A topic, then, \textit{belongs} to a document in a given proportion. Our analysis relies on the probability computed the other way around: documents \textit{belonging} to a given topic.

To assign a document as belonging to a given topic we rely on the probabilistic notion of document likelihood. Once we choose a topic, a document is defined to be likely to belong to said topic if (1) the document itself is probable, i.e., the exact combination of words is likely to occur, and (2) the document contains said topic in large proportion.

Assuming that all documents are equally likely simplifies the computation of the likelihood to only topic proportions.\footnote{Mathematically speaking, we are using Bayes' rule to compute the probability of documents conditioned on topics. We are computing the \textit{maximum a posteriori} using a uniform prior over documents.} For each topic, we sort all articles according to the proportion of said topic in each article. This process reorganizes the corpus such that the ones at the top of the list discuss the chosen topic in great proportion. From this list, we consider the documents that cover 50\% of the total sum (that is, adding all proportions, and selecting the top documents in the list until we reach half of this sum).

Documents are thus assigned to each topic until we account for 50\% of the total likelihood sum, ordered from most likely to least likely relative to other documents in the corpus. Through this process of assigning documents to topics, a given document might be assigned to more than one topic.

There are two ad hoc choices in this process: we could have chosen another arbitrary percentage (besides 50\%), as well as another prior over documents (i.e., not assuming that all documents are equally likely). Different choices might yield better assignments by lowering the likelihood of documents that are rare in this context (e.g., literary pieces). Determining neutral criteria for such decisions is left for future research.

\section{Results}

\subsection{Main areas}

As explained above, we first assigned each topic to one of five main areas following the taxonomy used in the PhilPapers archive, plus one category for various articles which do not fall under any of the first five areas. Table \ref{tab:main_area_words} summarizes the words with the highest probability in each main area after averaging across time slices. These word lists are obtained after assigning topics to main areas based on each topic's word distribution. Therefore, obtaining a sensible word list for the main area validates the assignment of topics to specific main areas.

\begin{table*}[t]
\caption{Ten most probable words across all topics assigned to each main area.}\label{tab:main_area_words}
\begin{tabularx}{\textwidth}{l>{\raggedright}X}
\toprule
Main area & Most probable words \\ \midrule
Value Theory & moral, principio, experiencia, vida, hombre, libertad, juicio, humano, acción, razón \\[1.5em]
History of Western Philosophy & Rorty, Fichte, infinito, Hegel, Spinoza, pensar, Kant, Husserl, objeto, pensamiento \\[1.5em]
Science, Logic, and Mathematics & priori, ciencia, proposición, lógico, conocimiento, Cassirer, teoría, sistema, científico, geometría \\[1.5em]
Philosophical Traditions & educación, analogía, analítico, sujeto, analógico, pedagógico, Beuchot, aprendizaje, formación, educativo \\[1.5em]
Metaphysics and Epistemology & cuerpo, mundo, oración, dios, argumento, significado, justificación, escepticismo, escéptico, religioso \\[1.5em]
Other & filosofía, Mill, Socrates, Mundo, forma, Plato, Singer, self, relación, palabra \\ \bottomrule
\end{tabularx}
\end{table*}

We noticed several specific interests already emerging in these word lists. For ``History of Western Philosophy'', we observe an emphasis on several specific authors in the traditions of German idealism (Kant, Fichte, and Hegel) as well as interest in pragmatism (Rorty), phenomenology (Husserl), and early modern philosophy (Spinoza). For ``Philosophical Traditions'', there are several words related to philosophy of education (``educación'', ``pedagógico'', ``aprendizaje'', ``formación'', and ``educativo''), as well as an interest in the work of Mexican philosopher Mauricio Beuchot. Lastly, in ``Metaphysics and Epistemology'', there is a notable interest in religion (``dios'', ``religioso'') besides the expected distribution of words relating to general epistemology. 

\begin{figure*}[t]
    \centering
    \caption{Number of articles for each main area in the PhilPapers taxonomy in absolute quantity (left) and ratio of papers (right) in each area per year between 1951 and January 2022. Since an article may belong to more than one topic, it can also belong to more than one main area. Here we ignore overlaps between main areas, which means that the sum of documents in the graph is greater than the total number of documents in our corpus. We also include a line for 1983, the year in which the journal started publishing continuously after a period of hiatus.}
    \label{fig:num_articles_mainarea}
    \includegraphics[width=\textwidth]{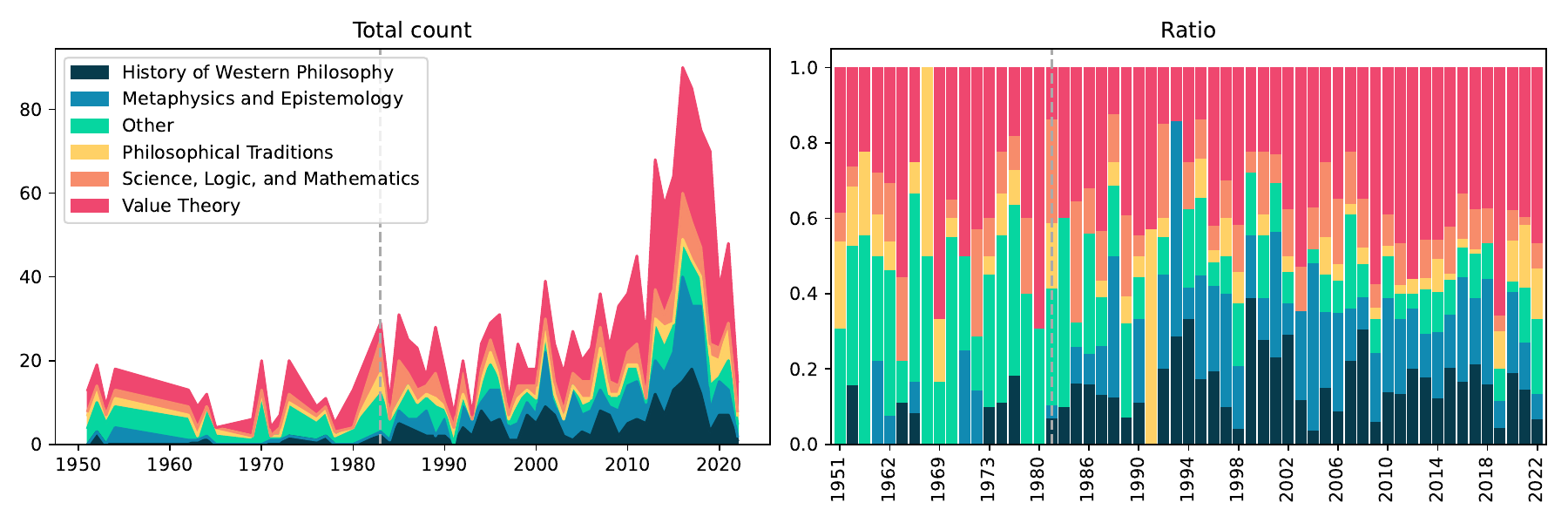}
\end{figure*}

Figure \ref{fig:num_articles_mainarea} shows the distribution of articles per main area, both in absolute terms and in proportion to the number of documents per year. The main area with the most documents overall is ``Value Theory'', with 571 documents assigned, followed by ``Metaphysics and Epistemology'' with 240 documents, ``History of Western Philosophy'' with 222 documents, ``Science, Logic, and Mathematics'' with 160 documents, and ``Philosophical Traditions'' with 84 documents.\footnote{Notice that a document may belong to more than one main area (thus explaining why these numbers sum to more than the number of documents in our corpus, 875). This is due to our criteria for determining whether a document \textit{belongs} to a given topic. See Sec.~\ref{subsec:assigning_docs_to_topics}.}  There were 231 documents classified as ``Other'', which included letters and literary pieces that were part of the journal's scope during the 1970s.

We see an increase in publications starting from the 1990s and exploding in the 2010s. This also allows certain topic areas to appear more forcefully. This is most salient in the case of ``Metaphysics and Epistemology''. Before 1990, only 22 out of 216 documents contained topics assigned to this main area, covering only 10.2\% of publications. After 1990 we find 218 out of 659 documents, covering 33.1\% of publications. The main area ``Value Theory'' also saw a similar increase, with 118 out of 216 documents containing topics in this main area before 1990 (54.6\% of publications), and 453 out of 659 documents after 1990 (68.7\% of publications). Similarly, ``History of Western Philosophy'' covered 13.4\% of documents (29 of 216) before 1990 and 29.3\% (193 of 659) after 1990. Both ``Philosophical Traditions'' and ``Science, Logic, and Mathematics'' decreased in proportional size, however. ``Philosophical Traditions'' covered 13\% (28 of 216) of documents before 1990, but only 8.5\% (56 of 659) afterward, while ``Science, Logic, and Mathematics'' covered 22.2\% of documents (48 of 216) before 1990, and 17.0\% thereafter.\footnote{Notice again that there may be overlap between main areas due to how documents are assigned to topics. This explains why the ratio of documents containing topics in each main area for each main area does not sum to 100\%.}

\subsection{Subareas}

\begin{figure*}[t]
    \centering
    \caption{Proportion of articles per year for each main area and largest subarea within each main area.}
    \label{fig:subareas}
    \includegraphics[width=\textwidth]{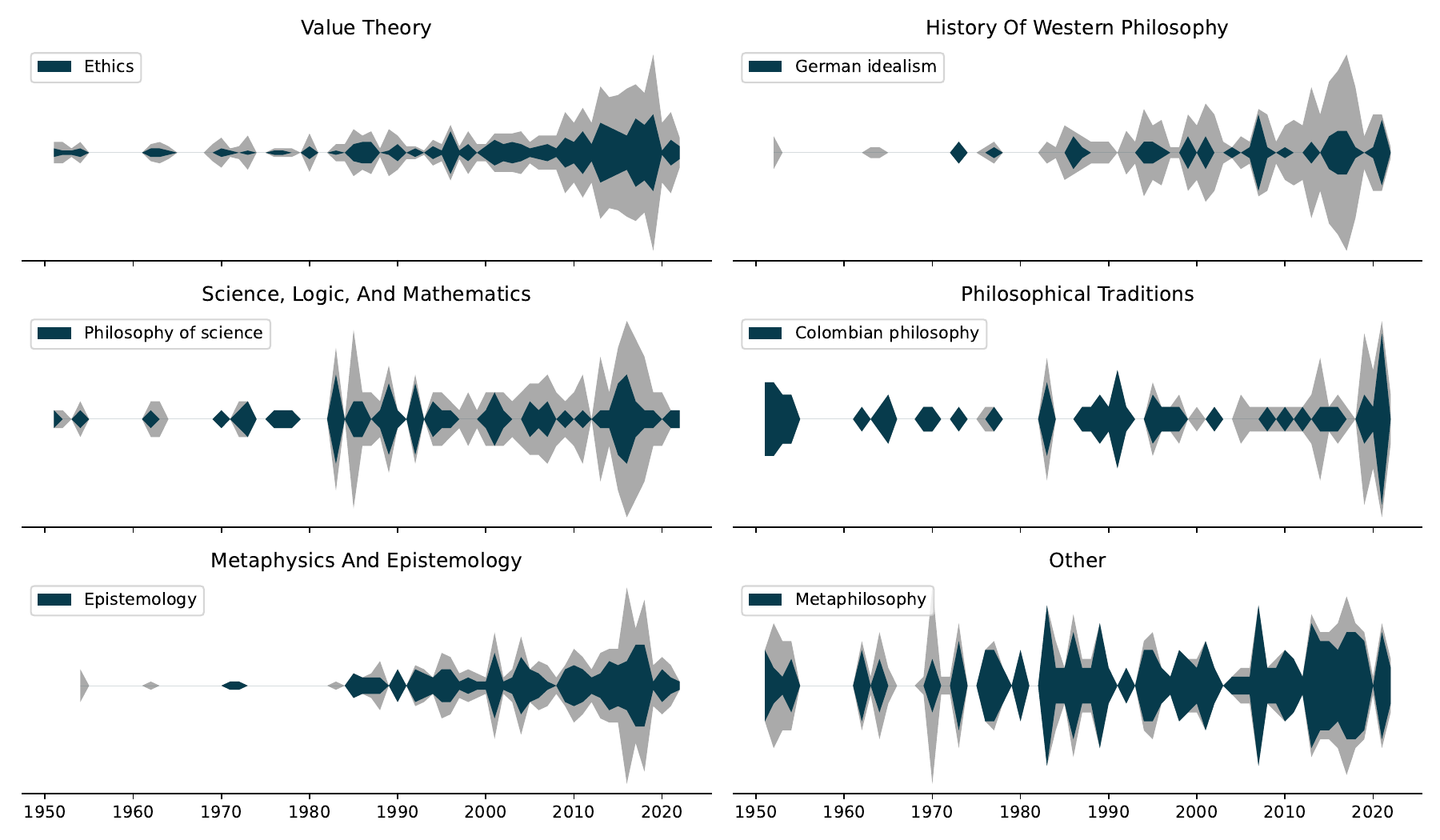}
\end{figure*}

For each of the main areas, we can also obtain the subareas with the most publications in that main area to get a better picture of which specific areas are more representative. Table \ref{tab:subareas} shows the number of documents for the five subareas with the most documents in each main area. For each of these subareas, we compute the ten most probable words, aggregating across all topics assigned to the corresponding subarea. Figure \ref{fig:subareas} shows the volume of the largest subarea within each main area through time.

\begin{sidewaystable}
\caption{Number of articles and most probable words for each subarea within each main area.} \label{tab:subareas}
\begin{tabularx}{\textwidth}{p{2cm}p{4cm}p{10cm}l}
 \toprule Main area & Subarea & Most probable words & Num. of docs \\ \midrule
Value Theory & Ethics & principio; vida; capacidad; enfoque; razón; discurso; idea; teoría; ética; desarrollo & 221 \\
     & Political theory  &                        principio; capacidad; enfoque; idea; desarrollo; teoría; humano; Sen; Nussbaum; noción &    110\\
     & Social and political philosophy &                   político; política; modelo; social; resultado; sujeto; Arendt; factor; Marx; condición &     72 \\
     & Kant  &                             libertad; juicio; voluntad; libre; Hume; bello; gusto; derecho; estético; natural & 68 \\
     & Theories of justice & principio; idea; teoría; noción; bien; original; primario; racional; Rawls; plan & 59 \\
\multirow{2}{=}{History of Western Philosophy} & German idealism &                Fichte; Hegel; pensar; pensamiento; lógica; conciencia; espíritu; absoluto; negación; posición  &     50 \\
      & Hegel &            Hegel; pensar; pensamiento; lógica; conciencia; espíritu; absoluto; contenido; Hegeliano; negación  &     48\\
      & Kant &               Kant; razón; kantiano; crítica; conocimiento; ilustración; naturaleza; práctico; concepto; puro  &     47\\
      & Phenomenology &                  Husserl; objeto; Heidegger; ente; mundo; propiedad; acto; referencia; conciencia; percepción &     46 \\
      & Epistemology  &         Kant; concepto; objeto; trascendental; intuición; conocimiento; unidad; representación; puro; espacio &     33\\
\multirow{2}{=}{Science, Logic, and Mathematics} & Philosophy of science  &         priori; ciencia; conocimiento; Cassirer; científico; geometría; teoría; constitutivo; problema; grupo &     62\\
      & Philosophy of language  &              proposición; lógico; lenguaje; sujeto; predicado; lengua; semántico; lógica; lingüístico; formal &     33\\
      & Logic &                    proposición; lógico; sujeto; lógica; predicado; necesario; formal; falso; enunciado; modal &     32 \\
      & Philosophy of physical science &                            física; ley; físico; espacio; teoría; sistema; movimiento; cuerpo; natural; Newton   &     22 \\
      & Truth  &  proposición; lógico; sujeto; predicado; lógica; necesario; formal; falso; modal; contingente &     22\\
\multirow{2}{=}{Metaphysics and Epistemology} & Epistemology  &  argumento; justificación; justificar; razón; normativo; posición; justificado; sostener; objeción; contenido &    126\\
      & Relativism  & creencia; problema; caso; tipo; teoría; término; mente; criterio; concepto; considerar &    102\\
      & Philosophy of religion  &                 dios; religioso; religión; teología; cristiano; divino; razón; Levinás; cristianismo; palabra &     37 \\
      & Realism &         mundo; realidad; real; realismo; conceptual; metafísico; contenido; realista; representación; ficción &     27 \\
      & Theology  &                            dios; teología; divino; cristiano; Levinás; palabra; razón; rostro; paz; teológico &     23\\
      Philosophical Traditions & Colombian philosophy  &                             filosofía; siglo; obra; historia; año; cultura; nacional; estudio; autor; revista &     53\\
      & Derrida / Ricoeur / Philosophy of Language*  &                     signo; texto; escritura; lenguaje; Derrida; palabra; Ricoeur; discurso; literatura; parís &     24\\
      & Latin American philosophy  &     educación; analogía; analítico; analógico; sujeto; pedagógico; Beuchot; aprendizaje; formación; educativo &      7\\
Other & Metaphilosophy / Phenomenology*  &              filosofía; concepto; relación; historia; forma; objeto; mundo; filosófico; realidad; pensamiento &    170\\
      & Literary pieces &                    palabra; amor; mundo; hombre; obra; mito; forma; dejar; vida; poesía  &     57\\
      & English  & Mill; Socrates; Plato; singer; self; political; philosophy; like; world; education &      4 \\ \bottomrule
\end{tabularx}
\footnotetext{Note: Subareas marked with `*' aggregate different subareas which we detected contained exactly the same number of items and most probable words. This is likely due to there being one single topic tagged with these subareas.}
\end{sidewaystable}

Starting with the largest main area, ``Value Theory'', we see a clear interest in topics related to ethics and political philosophy. This is notable because under the PhilPapers taxonomy, ``Value Theory'' also covers topics related to aesthetics. Hence, it appears that publications in ethics and political philosophy outweigh publications in aesthetics. Another interesting finding is that there is special interest in Kantian practical philosophy, given that ``Kant'' appears as one of the largest subareas within ``Value Theory''. Other specific subareas within ``Value Theory'' include Nussbaum and Sen's capability approach and theories of justice. Lastly, interest in ``Ethics'', the largest subarea within ``Value Theory'', has remained proportionally constant relative to publications in its main area. As shown in Fig.~\ref{fig:subareas}, ``Ethics'' scales at a similar rate to its corresponding main area.

Regarding ``History of Western Philosophy'', we see a clear interest in German philosophy: German idealism, Hegel, Kant (both general Kantian scholarship and Kantian epistemology), and Phenomenology (specifically Husserl and Heidegger). Its largest subarea, ``German idealism'', has constant publications through time, but these do not grow proportionally as the corresponding main area grows. This suggests that as publications in ``History of Western Philosophy'' increase, different subareas appear within this topic beyond publications on German idealism. We also find it noteworthy that the four largest subareas within ``History of Western Philosophy'' (those presented in Table \ref{tab:subareas} excluding ``Epistemology'') have relatively similar publication numbers, ranging from 46 to 50 items in each subarea. This means that interest in ``History of Western Philosophy'', while focused on German philosophy, remains somewhat balanced between German idealism, Hegel, Kant, and German Phenomenology.

On the side of ``Science, Logic, and Mathematics'', we observe a focus on philosophy of science (especially in topics related to physics) and philosophy of language and logic. In contrast to the previous main area, here we note a distinct interest in ``Philosophy of science'', which almost doubles the number of items assigned to the second-largest subarea, ``Philosophy of language.'' When considering its evolution through time, we observe constant interest throughout the years, but we also see an increase in publications beyond ``Philosophy of science'' in the last decade. We take this to imply a growing diversity of topics as the number of publications grows, similar to what we observe in other areas.

Fourth, concerning ``Metaphysics and Epistemology'', we observe two clear areas of interest: epistemology (including relativism and realism) and philosophy of religion. The first subarea, epistemology, substantially outweighs philosophy of religion, with almost 100 more items than the latter. Nevertheless, interest in philosophy of religion and theology is not negligible, accounting for more items than subareas in other main areas such as ``Philosophy of language.'' We also note that in both ``Philosophy of religion'' and ``Theology'', one author who stands out is Levinas, whose work has influenced other fields such as ethics and political philosophy. This suggests that part of the interest in these latter topics stems from a strong focus on these subareas, as discussed previously. In terms of time dynamics, ``Epistemology'' sees a proportional growth relative to the whole main area, similar to ``Ethics'' in ``Value Theory'' and contrasting with the largest subareas within ``History of Western Philosophy'' and ``Science, Logic, and Mathematics.''

In ``Philosophical Traditions'' we see two broad areas of interest: Colombian and Latin American philosophy and post-structuralist philosophy of language (especially Derrida and Ricoeur). We note that concerns with Colombian philosophy have been constant throughout the journal's history and particularly salient in the first decades. This is consistent with the journal's policies and early discussions toward a unique and localized style of philosophy and publication.

Lastly, in the ``Other'' category, we see a large concentration of articles under the ``Metaphilosophy / Phenomenology'' aggregated subarea. This topic covers a very heterogeneous set of publications which to some extent contain metaphilosophical reflections. These include discussions on phenomenological methods or Wittgenstein's metaphilosophy, for example. As such, we do not find these topics more informative than the other areas previously reported, since they cover almost every topic in the corpus. We also see some articles which correspond to ``Literary pieces'' published during the journal's first decades, before the journal shifted its publication policy toward academic articles. Finally, our model identified some English words as a single topic, mainly due to specific terms that were kept after eliminating general tokens in English as stopwords.

\subsection{Exegetical topics}

As explained before, \emph{Ideas y Valores} was founded with the aspiration of promoting a philosophical style geared toward original work rather than the exegesis of others' thought (Sec.~\ref{sec:history_of_the_journal}). This aspiration was echoed decades later by academics close to the journal who thought that philosophy in Colombia focused too much on the history of ideas rather than on pursuing its own problems and agendas \citep{IsazaDuque2013,SierraMejia1967}. We take it as motivation to ask whether exegetical topics became comparatively less prominent as this non-exegetical style of philosophy became more consolidated within Colombian and Latin American academia, a process various historians of philosophy in the region describe as the \emph{normalización} of the discipline \citep{Jimenez2012,Gomez2021,Lopez2021,Betancur2015a}. This gives us a hypothesis that we can partially test using our model.

To do this, we assigned special tags to topics that engaged specifically with historical ideas (e.g., dealing with the interpretation of a specific philosopher or work rather than a direct discussion building on their ideas). Table \ref{tab:historical_topics} shows the topics with the most documents per main area along with their assigned subareas, most probable words, and number of documents assigned. Taking the whole corpus into consideration, 27.78\% of documents were assigned to at least one of these topics.

As shown in Table \ref{tab:historical_topics}, the topics with most assigned documents among exegetical topics are those engaging with German philosophers (Kant, Hegel, Husserl, Heidegger). This is consistent with the emphasis on German idealism and phenomenology reported earlier. Other important figures that can be noticed in our analysis include Aristotle, Deleuze, Quine and Davidson (both authors within one topic), and Derrida and Ricoeur (also both authors within one topic). Furthermore, we found other topics that were specific to discussions of specific philosophers but only contained a handful of assigned documents. These include topics specialized in Nietzsche (13 documents), Wittgenstein (12 documents), Gadamer (11 documents), Spinoza (9 documents), Descartes (6 documents), and Rorty (5 documents).

It is worth noting that several of these philosophers were still alive and active during part of the period covered by the journal. The most significant cases are Deleuze, Quine, Davidson, Derrida, Ricoeur, Gadamer, and Rorty. We manually inspected the topics containing these mentions to ensure that the topic's nature was exegetical and not a direct discussion of these philosophers as active figures in the corresponding debates. In all cases, the articles with the highest likelihoods corresponded to articles published years after these philosophers' retirement or passing, supporting the interpretation of these topics as focusing on more exegetical and historical work rather than participating in an ongoing discussion. Most of the articles included in these topics were published past the year 2000, with very few cases of articles before that (and mostly articles published in the late 1990s). This is even the case for topic 49, which is devoted to articles on Rorty and encompasses five articles published in 1999 (the only article contemporary with Rorty's academic work), 2008 (in a special issue on Rorty's legacy), and 2010.\footnote{The specific data on these topics is available in our GitHub repository, including the list of most likely articles in each topic and the list of topics tagged accordingly (see \nameref{sec:data_availability}).}

To account for this expected consolidation of a non-exegetical style of philosophy, we hypothesized that topics engaging specifically with historical ideas would decrease in size throughout the years. Concretely, we considered the ratio of documents assigned to exegetical topics against the total documents published per year, and hypothesized that there would be a negative correlation between this ratio and time. By considering this ratio, rather than the total number of documents per topic, we account for differences in the number of articles published across time, especially given that there is a considerable increase in publication after 2010. Also worth noting is that we compute this model for publications after 1983, since we do not have continuous data before this year due to various interruptions in the journal's publication (see Sec.~\ref{sec:history_of_the_journal}).

\begin{figure}[t]
    \centering
    \caption{Simple linear regression on the ratio of documents in exegetical topics to documents in non-exegetical topics against time by year, using a 95\% confidence interval.}
    \label{fig:historical_correlation}
\includegraphics[width=0.8\textwidth]{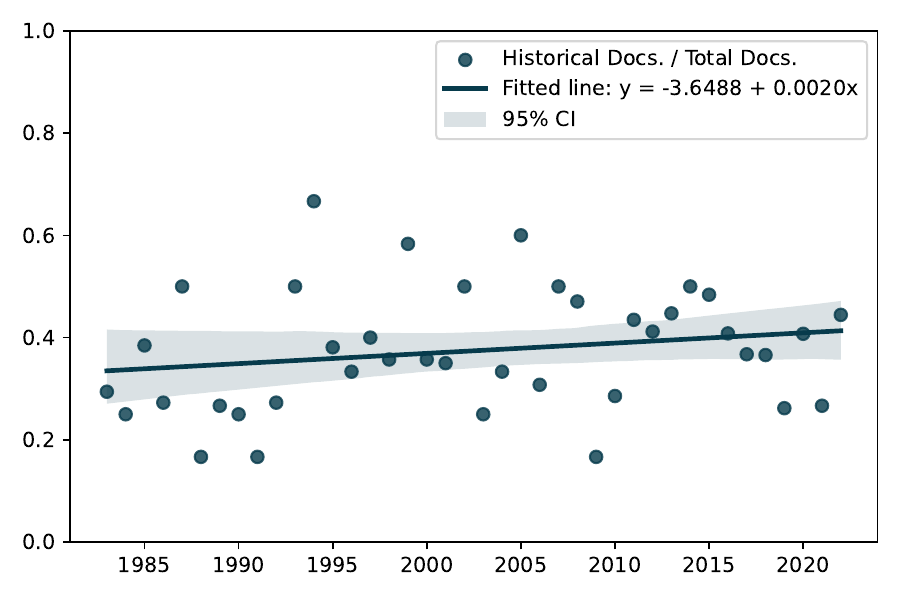}
\end{figure}

Figure \ref{fig:historical_correlation} shows a simple linear regression model fitted to the ratio of articles assigned to exegetical topics versus the total number of published articles across time. Computing Pearson's correlation coefficient with a one-sided test ($H_1: r < 0$) shows that the linear relationship between the ratio and time is statistically non-significant ($r(38) = 0.20, p = 0.8885$). Therefore, there is no evidence that the ratio of articles assigned to exegetical topics and the total number of articles per year has decreased through time. Moreover, visual inspection of the data suggests that there is effectively no change in this ratio through time, and we note that plotting confidence intervals at 95\% shows that there are models which explain the data both with a positive and a negative slope. Yet, we observe a slight tendency toward a positive slope, which suggests that there are more models which account for the data and have a positive slope. This is consistent with the calculated slope of 0.0020 in our model.

Given these non-significant correlation, we find no evidence that the ratio of articles in exegetical topics against those in non-exegetical topics has changed as academic philosophy in Colombia became more consolidated as a discipline. On the contrary, it appears more likely that this ratio has remained constant throughout the years covered by our analysis, despite the journal's founding aspiration toward more original and non-exegetical work. However, we are cautious about this conclusion insofar as it is based on a non-significant statistical analysis as well as on an interpretation of topics as mostly exegetical. We discuss some of these limitations below.

\begin{sidewaystable}
\caption{Summary of topics labeled as exegetical topics and to which more than five documents were assigned.}
\label{tab:historical_topics}
\begin{tabularx}{\textwidth}{
>{\raggedright}p{0.1\textwidth}
>{\raggedright}p{0.2\textwidth}
Xl}
\toprule
Main area & Subareas & Most probable words &   Num. of docs  \\
\midrule
\multirow{2}{=}{History of Western Philosophy} & Kant; Epistemology & Kant; concepto; objeto; trascendental; intuición; conocimiento; unidad; representación; puro; espacio &            33 \\
             & German Idealism; Hegel & pensar; pensamiento; lógica; negación; movimiento; contenido; Hegel; lógico; puro; singular &            32 \\
             & Phenomenology; Husserl & objeto; propiedad; referencia; relación; percepción; descripción; nombre; acto; intencional; causal &            18 \\
             & Greek Philosophy; Aristotle & Aristóteles; movimiento; naturaleza; aristotélico; alma; forma; Platón; natural; esclavo; pasaje &            17 \\
             & Phenomenology; Heidegger & Heidegger; ente; mundo; cosa; ontológico; Heideggeriano; relación; interpretación; dasein; existir &            17 \\
             & German Idealism; Hegel & Hegel; conciencia; espíritu; absoluto; Hegeliano; autoconciencia; fenomenología; reconocimiento; sujeto; concepto &            16 \\
             & Modern Philosophy; Kant; Political Theory & Kant; razón; kantiano; ilustración; crítica; práctico; naturaleza; prejuicio; biblia; hombre &            14 \\
             & Nietzsche & nietzsche; voluntad; fuerza; mundo; vida; forma; devenir; nihilismo; ideal; bataille &            13 \\
             & Analytic Philosophy; Wittgenstein & lenguaje; Wittgenstein; juego; regla; palabra; lógico; expresión; pensamiento; proposición; práctica &            12 \\
             & Hermeneutics; Gadamer & Gadamer; hermenéutico; hermenéutica; comprensión; interpretación; comprender; tradición; horizonte; lenguaje; diálogo &            11 \\
             & Phenomenology; Husserl & Husserl; conciencia; fenomenología; objeto; trascendental; fenomenológico; acto; intencional; Husserliano; vivencia &            11 \\
             & Modern Philosophy; Spinoza; Metaphysics & infinito; Spinoza; naturaleza; actividad; infinitud; acción; concepción; aspecto; implicar; determinado &             9 \\
             & Modern Philosophy; Descartes & Descartes; cartesiano; certeza; alma; pasión; meditación; vii; existencia; cogito; cuerpo &             6 \\
\multirow{2}{=}{Metaphysics and Epistemology} & Deleuze & síntesis; deleuze; estética; sensación; lakoff; pasivo; placer; deleuziano; displacer; contracción &             9 \\
             & Analytic Philosophy; Quine; Davidson & oración; significado; quine; davidson; evidencia; teoría; observacional; empirismo; sujeto; distal &             6 \\
Philosophical Traditions & Philosophy of Language; Derrida; Ricoeur & signo; texto; escritura; lenguaje; Derrida; palabra; Ricoeur; discurso; literatura; París &            24 \\
             & Latin American Philosophy; Hermeneutics; Beuchot & educación; analogía; analítico; analógico; sujeto; pedagógico; Beuchot; aprendizaje; formación; educativo &             7 \\
Value Theory & Philosophy of Law; Kant & derecho; jurídico; ley; natural; naturaleza; político; deber; Kant; hombre; voluntad &            29 \\
             & Social and Political Philosophy; Marx & Marx; social; relación; clase; capital; forma; materialismo; ideología; producción; sociedad &            20 \\
             & Aesthetics; Kant; Modern Philosophy & juicio; bello; gusto; estético; juzgar; belleza; común; objeto; resultar; obra &            17 \\
\bottomrule
\end{tabularx}
\end{sidewaystable}

\section{Discussion}

In this study, we modeled the evolution of topics in an established philosophy journal in Colombia and Latin America, \emph{Ideas y Valores}. We used a Dynamic Topic Model, a model which expands on Latent Dirichlet Allocation (LDA) models to also include temporal dynamics in the behavior of different topics. This allowed us to reconstruct a history of philosophy in Colombia (and perhaps part of the history of philosophy in Latin America) using digital methods.

Our findings show that throughout the journal's history, most articles have engaged with topics related to ``Value Theory'', particularly to ethics, political philosophy, and social and political theory. This main area has twice as many documents as the second most populated area (``Metaphysics and Epistemology''), suggesting a strong focus on these topics. As seen in Fig.~\ref{fig:num_articles_mainarea}, this area has also been a popular area throughout the journal's history, maintaining a strong presence proportional to the total articles published across the years. There are various possible explanations for this observation. First, many of the journal's founders had interests in political philosophy, specifically in the theory of law \citep{Tovar2000}. This shaped the journal's focus on these topics in its earlier years. Second, the country's history of political upheaval and conflict has led several scholars to call for a philosophical reflection on topics such as justice, violence, and forgiveness \citep{SierraMejia2002,HerreraRomero2022}. We encourage further research to explore the specific topics and evolution of concerns across Colombian history through the lens of these historical periods.

In the case of ``Metaphysics and Epistemology,'' we observed a strong focus on epistemology and relativism, which explains the strong presence of ``Rorty'' in ``History of Western Philosophy.''  Concerning this latter category, we see a special interest in Phenomenology and German Idealism. Lastly, regarding ``Science, Logic, and Mathematics'', we see discussions in philosophy of science and philosophy of language, areas which we interpret as a signal of the emergence of analytic philosophy within the journal. This interpretation is further supported by the fact that topics in these latter areas emphasize traditionally analytic discussions on logic, physics, and empiricism.

Besides analyzing areas and subareas throughout the corpus, we also investigated whether the journal has changed its historical focus throughout the years. We noticed that while some academics proposed steering away from historical discussions and promoting some form of originality, the journal has always included both historical and non-exegetical topics within its articles. However, these findings must be explored in more detail with different techniques, since our analysis relies on us tagging certain topics as historical or non-historical, tags which may be open to discussion. Future studies may approach these problems by using independent raters or odd-one-out techniques which may make classification more precise \citep{Chang2009}.

Another result worth noting is the emphasis on German philosophers, particularly Kant, Hegel, Heidegger, Husserl, Nietzsche, and Gadamer, among others. We note that German thinkers appear more prominently throughout the corpus than, for instance, French or US American philosophers. This is most clearly seen when comparing subareas tied to these traditions directly: the German-focused subareas within ``History of Western Philosophy'' (German idealism, Hegel, Kant, Phenomenology) total 191 documents, compared to only 24 in the ``Derrida / Ricoeur'' subarea representing French post-structuralist philosophy (Table \ref{tab:subareas}). This is further evidenced by the fact that we find more documents focusing on German philosophers than other groups, especially in the subareas included in ``History of Western Philosophy'', and that several of the most prominent words in various topics mention German philosophers specifically and more frequently than non-German philosophers (Table~\ref{tab:subareas}). Additionally, few non-German philosophers appear among the most probable words for each main area (Rorty, Spinoza, Beuchot) besides the ``Other'' main area  (which includes Mill, Socrates, Plato, and Singer, but collects heterogeneous and non-classifiable documents, and each of these words has a lower likelihood within each topic than other more significant words in other areas), yet we observe several German philosophers among the most probable words of the largest main areas (Fichte, Hegel, Kant, Husserl, and Cassirer) (Table~\ref{tab:main_area_words}). Such visibility of German thinkers could be explained by the fact that many Colombian philosophers (e.g., Jorge Aurelio Díaz, Guillermo Hoyos, Rubén Sierra Mejía) studied in Germany between 1960 and 1990 (and many continue to do so). There is also a strong and documented interest in phenomenology throughout the journal's history \citep{Mejia1988,Tovar2000}.

\subsection{Limitations}

Topic modeling techniques have become much more accessible in recent years with the advent of more user-friendly packages and technologies to implement models such as LDA. However, there are still a number of different workflows and decisions which have not been standardized either in the literature in digital humanities and philosophy or in general discussions on the use of topic models in computer science.

One limitation we encountered concerns selecting the appropriate number of topics for our model. While we opted for a relatively data-driven approach using coherence, log perplexity, and number of articles without a topic as relevant metrics, much more can be done to make the human decisions involved more explicit and reproducible. It is common to use coherence alone as a metric to select the number of topics for the model \citep[see e.g.,][]{Greene2024} or to rely on expert judgment to determine how interpretable the resulting topics are \citep[see e.g.,][]{Malaterre2021}. Instead of limiting ourselves to one metric, we decided on two additional metrics which, after manual inspection, offered a plausible model for our data. This combined both approaches, and while it improved on previous methods to make these decisions, we also think this step should be refined further in the literature. For example, some alternatives to this workflow could add word intrusion or topic intrusion tasks as suggested by \citet{Chang2009} to verify empirically that the selected model is the most interpretable.

A related limitation concerns the use of the PhilPapers taxonomy. While we opted for a more transparent, publicly accessible, and openly debatable taxonomy, there are still drawbacks in using a taxonomy based on anglophone philosophy to understand philosophy in Latin America. Many of the categories in this taxonomy are unbalanced in granularity relative to the concerns and specializations observed in our corpus. As we noted above, for example, categories related to phenomenology and hermeneutics are lumped together, while categories in epistemology and metaphysics get more granular distinctions. We believe that any taxonomy constructed from a specific philosophical background or intuition will be biased toward certain areas of interest and expertise, and therefore we decided to use a taxonomy that, while admittedly as biased as other alternatives, could be debated openly. We also sought a taxonomy that enabled contrasts with other locations and respected the journal's initial aspiration to develop a philosophical style that could be discussed universally. As such, accepting English as the dominant language of contemporary academic production, we believe that an anglophone taxonomy constitutes a better basis for comparison than a localized taxonomy. Yet, the construction of a global taxonomy for philosophy is precisely an area where data-driven approaches could be helpful, and future research could explore the construction and comparison of various philosophical taxonomies across the globe. In the case of Latin America, these taxonomies could also be constructed along with experts in Latin American history of philosophy, enabling further dialogue with Latin American philosophical historiography while avoiding risks of epistemic exoticization \citep{Davis2016,Valderrama2025}.

In terms of interpretive and philosophical resources, we lastly note that the distinction between exegetical and non-exegetical topics could be drastically improved. As it stands, our distinction relies on manual inspection and tagging based on our own expertise. Future studies could improve on this by approaching the problem in a more data-driven fashion. Additionally, some exegetical aspects remained unexplored, such as the presence of Ernst Cassirer as one of the most important names in the ``Science, Logic, and Mathematics'' main area. His appearance suggests that even within topics that are often interpreted as having a contemporary focus, such as philosophy of science and epistemology, there are also exegetical and historical discussions that may remain undetected.

In terms of technical resources, an important limitation of our study was due to differences in the resources available for topic modeling in Spanish. Even though NLP and topic modeling tools are widely available in English, the repertoire of tools for Spanish is much more restricted. These include stopword lists and workflows for orthographical correction, which are much smaller than their English-language counterparts. For the present study, we implemented several strategies to overcome these limitations, such as extending word lists available in NLTK and spaCy with other lists available online, as well as using a subset of our corpus to correct artifacts in OCR in Spanish. We expect that future developments of these tools in other languages will enable more diversity in the types of studies carried out in the fields of digital humanities, data-driven approaches to philosophy and its history, and more general applications of text mining tools. 

Lastly, there was information that we did not analyze and that could offer further insights into the history and present of philosophy in Colombia and Latin America. This includes information present in the metadata about authors' gender (as inferred from their names) and institutional affiliation. We think examining such metadata may lead to interesting findings about the demographic distribution within certain topics. Future studies could also exploit citation and reference information to study reference networks within and between topics and areas. While we did not carry out these analyses, we expect to explore these data in the future.

Overall, we see a promising area of study to apply models such as DTMs and LDAs to answer questions about the history of philosophy, both globally and locally. Despite the limitations of the present study, we take this as an opportunity to promote the use of digital technologies to shed light on how philosophical texts have evolved to engage with new ideas, styles, currents, and developments. We expect future studies to include other journals, both in Colombia and internationally, and to develop new tools for the analysis of such corpora.

\section*{Declarations}

\subsection*{Authorship declaration}

Both authors have contributed equally to all aspects of this work. Both authors have participated in the conception and design of the study, contributed to data collection, analysis, and interpretation, been involved in drafting and critically revising the manuscript, approved the final version of this manuscript, and agreed to be accountable for all aspects of the work.

\subsection*{Data and code availability}
\label{sec:data_availability}

The complete set of analyses, as well as the code base, the stopword lists, and dictionaries are available on GitHub: \url{https://github.com/juanrloaiza/latinamerican-philosophy-mining}.

\subsection*{Declaration of competing interests}

The authors declare that they have no competing interests relevant to the content of this article. No funding was received for conducting this study.

\subsection*{Acknowledgments}

We would like to thank Nicolás Duque-Buitrago for his incredibly helpful comments and support during the revision process of this article.

\bibliography{refs}

\end{document}